\pdfoutput=1
\documentclass[11pt]{article}

\usepackage{acl}

\usepackage{times}
\usepackage{latexsym}
\usepackage{multirow}
\usepackage{graphicx}
\usepackage{makecell}
\usepackage{booktabs}
\usepackage{tabularx}

\usepackage[T1]{fontenc}
\usepackage[utf8]{inputenc}

\usepackage{microtype}

\usepackage{inconsolata}
\usepackage{amsmath}
\usepackage[misc,geometry]{ifsym}

\title{Potential and Challenges of Model Editing for Social Debiasing}

\author{%
Jianhao Yan$^{1,2}$\thanks{~~These authors contributed equally to this work.} 
\hspace{1em}Futing Wang$^{1,2}$\footnotemark[1] 
\hspace{1em}Yafu Li$^{1,2}$ 
\hspace{1em}Yue Zhang$^{2,3,\text{\Letter}}$ \\
\centerline{\normalfont{$^1$Zhejiang University} \quad \normalfont{$^2$School of Engineering, Westlake University}} \\
\centerline{\normalfont{$^3$ Institute of Advanced Technology, Westlake Institute for Advanced Study}} \\
\centerline{\texttt{elliottyan37@gmail.com}}
}

\begin{document}
\maketitle
\begin{abstract}
  \textbf{Warning}: This paper contains content that is stereotypical and may be upsetting.
  
  Large language models (LLMs) trained on vast corpora suffer from inevitable stereotype biases. Mitigating these biases with fine-tuning could be both costly and data-hungry. Model editing methods, which focus on modifying LLMs in a post-hoc manner, are of great potential to address debiasing. 
  However, it lacks a comprehensive study that facilitates both internal and external model editing methods, supports various bias types, as well as understands the pros and cons of applying editing methods to stereotypical debiasing. 
  To mitigate this gap, we carefully formulate social debiasing into an editing problem and benchmark seven existing model editing algorithms on stereotypical debiasing, i.e., debias editing. 
  Our findings in three scenarios reveal both the potential and challenges of debias editing:
  (1) Existing model editing methods can effectively preserve knowledge and mitigate biases, while the generalization of debias effect from edited sentences to semantically equivalent sentences is limited. 
  (2) Sequential editing highlights the robustness of SERAC~\cite{serac}, while internal editing methods degenerate with the number of edits. 
  (3) Model editing algorithms achieve generalization towards unseen biases both within the same type and from different types.
  In light of these findings, we further propose two simple but effective methods to improve debias editing, and experimentally show the effectiveness of the proposed methods. 
  % a benchmark that suits all existing model editing methods
  % There lacks a benchmark that suits all existing model editing methods and a systematical study that reveals the pros and cons of editing social biases on LLMs. 
  % In this paper, we systematically study editing methods for social debiasing in large language models. 
  % To this end, with carefully formulating the social debiasing into an editing problem, we first construct a benchmark dataset based on pair-wise comparison data, and propose an evaluation protocol that comprehensively evaluates the edited model in both single and sequential edits. 
  % Furthermore, we 
  % We show that, with the state-of-the-art MEMIT algorithm, we successfully reduce the social bias of language models while maintaining their performance in other tasks.
  % Specifically, we first design a synthetic dataset that encodes social biases including racial bias, gender bias, and stereotypes. We apply the locating algorithm and find the biases are mainly encoded in the MLP layers of entities.
  % By designing a de-bias objective that minimizes the gap between uniform distribution and the distribution across synonyms and antonyms of biased words, we wipe out the social bias.
  % We evaluate our method on various language models and bias benchmarks, and show that with only a few bias samples, the edited models well generalize to unseen bias samples and bias types.
  % Our results suggest the potential of extending model editing methods to various aspects.  
\end{abstract}

\section{Introduction}

% \section{Introduction}

% Large Language Models (LLMs) like GPT-4 and LLaMA2 have shown significant advancements in natural language understanding and task execution. The effectiveness of LLMs is attributed to the scaling law~\cite{kaplan2020scaling,hoffmann2022training} which posits that larger models yield better performance. However, this scaling complicates and increases the cost of model fine-tuning for knowledge updates and error correction. Model editing has emerged as a promising solution for modifying pre-trained models' outputs~\cite{meng2022locating,meng2023massediting,zhang2024comprehensive}.

Large Language Models (LLMs) perform excellently across various downstream tasks, allowing a wide range of applications, ranging from chatbots to medical diagnoses \cite{wang2023chatcad} to robotics \cite{driess2023palme}. However, LLMs trained on vast corpora can inadvertently learn biased information, leading to
negative stereotypes and social biases encoded within the models \cite{gallegos2023bias}. For instance, when given \textit{'Arab people are rich.'} and \textit{'Arab people are poor.'}, the LLaMA-2-7B~\cite{llama-2} model favors the first sentence, which reflects stereotypes encoded within the model. Such biases have the potential to result in unfairness and harm when deployed in production systems \cite{nadeem-etal-2021-stereoset, prakash-lee-2023-layered}. 

% More extensive challenges are associated with auditing and mitigating bias in Large Language Models (LLMs). Biases can manifest in intricate semantic, pragmatic, and commonsense knowledge~\cite{zhang2024comprehensive, wang2024decodingtrust}. On the other hand, 
Traditional debiasing methods typically require training models from scratch to modify training datasets~\cite{zmigrod-etal-2019-counterfactual, dinan-etal-2020-queens, qian-etal-2022-perturbation,narayanan-venkit-etal-2023-nationality}, optimization process \cite{huang2020reducing,qian2022perturbation,he2022controlling,park2023never,zhou2023causal}
% \cite{huang2020reducing,qian2022perturbation,he2022controlling,guo2022auto， park2023never,zhou2023causal,li2023prompt,yu2023unlearning}, 
or they may operate within the output space \cite{he-etal-2021-detect-perturb, tokpo-calders-2022-text, majumder2022interfair,dhingra2023queer}. The former is very costly for language models, while the latter does not truly address bias encoded in the model, potentially leading to non-robustness.

When it comes to mitigating specific biases in pre-trained models, such as the association between "nurse" and "women", directly fine-tuning~\cite{ghanbarzadeh-etal-2023-gender} the model might be both costly and impractical with insufficient data. 
On the other hand, model editing~\cite{mitchell2022fast,zhang2024comprehensive}, capable of post-training adjustments to the knowledge within the model, has shown potential in tackling this issue.
It is initially adopted for modifying pre-trained models' knowledge~\cite{meng2022locating,meng2023massediting,zhang2024comprehensive}.
% Existing methods can be categorized into intrinsic and external methods, where the former modifies the LLMs' parameters and the latter 
% Extensive studies have been devoted to analyze 
Recently, researchers have started to employ model editing methods for other tasks such as editing personality~\cite{edit-personality} or reasoning process~\cite{akyurek-etal-2023-dune}. 
As for editing stereotypical biases, there are several important research questions left unsolved: 
(1) Given the rich and complex semantics of stereotype sentences compared with factual knowledge, \emph{how to define the editing problem for debiasing?}
(2) \emph{What are the advantages and disadvantages of employing model editing methods to debiasing?}
(3) \emph{Can edits of debiasing generalize to unseen biases?}

In this work, we present a comprehensive study of the problem of model editing for social debias. We adopt a flexible formulation that takes the tokens before the first appearance of the subject as the prompt and the rest as our target. 
This formulation supports different bias types and various internal and external editing methods. Based on the formulation, we construct a debias editing dataset based on StereoSet~\cite{stereoset} to support both internal and external editing algorithms. 
By experimenting with seven editing algorithms over both LLaMA2-7B~\cite{llama-2} and GPT2-XL~\cite{gpt2}, we have the following observations: 
\begin{itemize}
  \item In the setting of single-edit, i.e., edit out a single biased sentence, model editing methods can achieve almost 100\% edit success rates and do not hurt LLMs' internal knowledge~(relatively high scores of Knowledge ACC). Nonetheless, generalization is challenging for debias editing. The edited sentence achieves a higher probability than the biased one as well as its paraphrases, but this effect hardly generalizes to the paraphrases of the edited sentence. 
  \item For sequential editing, i.e., edit out biased sentences one by one, all methods degenerate when increasing the number of edits, except for SERAC~\cite{serac} which performs consistently well when increasing edits. 
  With more edits, the complex structure of biased sentences brings a severe challenge to both debias success rate and hurt internal knowledge. 
  % suggesting memory selection and augmented outputs are the promising directions for debias editing. 
  \item We show that model editing algorithms can achieve generalization over bias types. After being edited on one specific bias type, e.g., Race, LLMs demonstrate some debiasing effect on another bias type, e.g., Profession.
\end{itemize}
% Furthermore, we identify two challenges for debias editing. The first one is to accurately select the scope for editing, including targets and parameters. 
% The second one is to improve the generalization of the edit effect from the edited sentence to its paraphrases. 
In light of our observations, we propose two simple methods to mitigate the challenge brought by many edits. 
We adopt a heuristic rule-based target selection and a causal tracing selection to constrain the scope of the edits.
Experimental results show that our methods demonstrate strong performance while retaining the edit performance. 
To the best of our knowledge, we are the first to comprehensively study the problem of debias editing, reveal its potential and challenges, and propose effective solutions. 
We release our dataset and codes to facilitate future work\footnote{\url{https://github.com/ElliottYan/ModelEditingForDebias}}.

\section{Related Work}
\paragraph*{Debiasing}
Addressing social bias in language models is an ongoing challenge that has received significant attention. Strategies for mitigating bias in language models can be classified based on different stages of the model workflow: Preprocessing techniques aim to detect and eliminate bias and unfairness early on, either within the dataset \cite{zmigrod-etal-2019-counterfactual, dinan-etal-2020-queens, abid2021persistent, qian-etal-2022-perturbation,ghanbarzadeh-etal-2023-gender} %yu2023mixup,thakur2023language,orgad2023blind,sun2023moraldial,omrani2023social，narayanan-venkit-etal-2023-nationality
or 
prompt \cite{mattern2022understanding, fatemi2023improving, yang2023adept}. In-training bias mitigation techniques focus on reducing bias and unfairness during model training, by adjusting model architecture \cite{bartl2020unmasking,han2022balancing}, modifying loss functions \cite{liu2020gender, webster2020measuring, ouyang2022training,woo2023compensatory, park2023never,zhou2023causal,li2023prompt}, or selectively updating parameters \cite{qian2022perturbation,ranaldi2023trip, yu2023unlearning}. Intraprocessing approaches alter decoding behavior \cite{saunders-etal-2022-first, meade2023using, kim-etal-2023-critic,chung-etal-2023-increasing, hallinan-etal-2023-detoxifying} without additional training or fine-tuning. Post-processing techniques primarily adjust model outputs to address bias and unfairness, without directly accessing the model itself \cite{he-etal-2021-detect-perturb,tokpo-calders-2022-text,  majumder2022interfair,dhingra2023queer}. However, effectively modifying bias in pre-trained large language models while minimizing disruption to the model's capabilities remains largely unexplored.

\paragraph*{Model Editing}
To address inaccuracies and biases in Large Language Models, various model editing techniques have been developed for efficient post-training adjustments. These can be categorized into two main types: intrinsic methods that modify the model's architecture or parameters, and extrinsic methods that adjust the input or output space \cite{akyurek-etal-2023-dune, zhang2024comprehensive}. Intrinsic editing involves direct changes to the model, such as parameter updates or new connections. For instance, simple fine-tuning to the model's original objectives is common, but it can lead to overfitting and forgetting previously learned information \cite{mitchell2022fast}. Alternative strategies involve editing model activations \cite{meng2022locating, meng2023massediting}, training auxiliary models to predict parameters \cite{de-cao-etal-2021-editing, tan2024massive}, or directly altering model representations \cite{hernandez2023inspecting} to incorporate new information. Some techniques focus on updating specific knowledge, while others use an external memory to store and retrieve edits. SERAC \cite{serac}, utilizes a scope classifier to determine the relevance of an edit and a counterfactual model to apply it, both of which require training for new edits.

% \section{Related Work}

\paragraph*{Model Editing for Debiasing}
Very recently, DAMA ~\cite{limisiewicz2024debiasing} identifies the stereotype representation subspace and edits bias-vulnerable FFNs using an orthogonal projection matrix. They propose a smart approach that uses profession as the subject and `he' or `she' as the target to facilitate causal tracing. However, their bias is limited to gender, making it difficult to generalize to other types of biases or relationships. 
% In contrast, our work adopt a flexible formulation that 
By extending our research scope to encompass four major categories of bias, we achieve more flexible debiasing strategies and comprehensively study the debias editing problem. 
% \elliott{discuss why DAMA is not editing, cannot support single edit. That's also why we do not compare them.} 
Besides, DUNE~\cite{akyurek-etal-2023-dune} broadens the scope of model editing to free-form natural language to involve editing with bias. This approach does not support intrinsic editing methods and is only suitable for external manner while we have explored both intrinsic and external model editing approaches.

\section{Experimental Settings}
In this section, we first describe our data collection and problem definition for debias editing. 

\subsection{Model Editing}
In literature, model editing is mainly applied to knowledge. 
The purpose is to modify $\mathcal{M} \rightarrow \mathcal{M}'$, overriding the undesired knowledge with a new one and keeping other knowledge intact. 
Specifically, given a prompt $p$, an editing algorithm $F$ edits the language model's next token prediction from knowledge $k$ to $k'$, 
\begin{gather*}
    k = \text{argmax}_{y}(P_{\mathcal{M}}(y|p)), \\
    \mathcal{M}' = F(\mathcal{M}, k, k'), 
\end{gather*}
where we refer to $k'$ as the \emph{target}. 
% The goal is to make sure the edited model $\mathcal{M}'$ overriding the undesired knowledge $k$ and keeping other knowledge intact,
Then, a successful edit should satisfy the following conditions,
\begin{gather*}
    k' = \text{argmax}_{y}(P_{\mathcal{M}'}(y|p)), \\
    \forall \hat{p}, \text{argmax}_{y}(P_{\mathcal{M}'}(y|\hat{p})) = \text{argmax}_{y}(P_{\mathcal{M}}(y|\hat{p})).
\end{gather*}
We refer to the acceptance of the first condition as \emph{edit success rate} and the second one as \emph{Knowledge Acc}. 

% \paragraph{Editing Algorithms}
We benchmark seven model editing algorithms, as listed below.
These methods can be divided into two categories, internal editing methods and external editing methods~\cite{zhang2024comprehensive}. Internal methods edit the model's parameters, applying techniques like causal tracing or meta-learning to find the specific module that contains key information. On the other hand, external methods keep the model's parameters untouched and utilize the external prompt/memory to change behavior. 

\begin{itemize}
  \item \textbf{FT}~(Internal) applies Adam with early stopping at one layer to minimize $-\log P(k'| p)$ following \cite{meng2023locating}, and the layer is chosen by causal mediation as in ROME.
  \item \textbf{FT-L}~(Internal) refers to Constrained Fine-Tuning \cite{zhu2020modifying}, which  additionally imposes a parameter-space $L_\infty$ norm constraint on weight changes. Note that FT and FT-L are different from direct fine-tuning and the optimization is performed edit by edit. 
  \item \textbf{MEND}~(Internal,~\citealt{mitchell2022fast}) applies the rank-one decomposition to divide the model into two rank-one matrices, from which it is possible to compute the $\Delta W$, significantly reducing the number of parameters. 
  \item \textbf{ROME}~(Internal, ~\citealt{meng2022locating}) employs a casual analysis method to detect which part of hidden states plays more importance. They view the editing as a minimum optimization and edit the weights. 
  \item \textbf{MEMIT}~(Internal, ~\citealt{meng2023massediting}) spreads updates evenly over the range of target layers to improve the robustness when parameter change magnitudes are minimized, while ROME modifies on a single layer.
  \item \textbf{SERAC}~(External, \citealt{serac}) stores input-output pairs into memory and retrieves a relevant edit using a learned scope classifier followed by a counterfactual model which is used in lieu of the main model. Both modules i.e. the scope classifier that identifies if an edit is relevant to the test query and the
counterfactual model. 
  \item \textbf{IKE}~(External, \citealt{zheng2023edit}) constructs
three types of demonstrations – copy, update, and retain – to aid the model in producing reliable fact
editing. It utilizes a demonstration store, formed from training sets, to guide the model toward
generating the appropriate answer by retrieving the most pertinent demonstrations. 
\end{itemize}

% \paragraph{Models and Implementation}
We use both LLaMA-2 and GPT2-XL as our base model. We present the results for LLaMA-2 and we put our results for GPT2-XL in the Appendix. 
We conduct our experiments based on EasyEdit~\cite{wang2023easyedit}. 

% \subsection{Debiasing with Model Editing}
\subsection{Debiasing Data Preparation}
Following previous work in debiasing~\cite{crows-pairs}, we define the debiasing problem with pairs of biased and unbiased sentences. 
Consider a pair of sentences $(x_{\text{more}}, x_{\text{less}})$, where $x_{\text{more}}$ is more stereotypical biased than $x_{\text{less}}$. We say a language model $\mathcal{M}$ is biased in this pair if the likelihood of $\mathcal{M}$ leans towards the more biased sentence, 
\begin{gather}
    P_{\mathcal{M}}(x_{\text{more}}) > P_{\mathcal{M}}(x_{\text{less}}).
\label{eq:bias}
\end{gather}
Thus, there are two ways towards debiasing with editing, either reducing the likelihood of $x_{\text{more}}$ or enhancing the likelihood of $x_{\text{less}}$.
In our experiments, we mainly focus on the latter way.
% A key difference from knowledge editing to bias editing is that stereotypically biased sentences generally do not contain a specific knowledge to edit. 
One key problem for applying model editing in debiasing is the definition of \emph{target}. Unlike knowledge editing, where the \emph{target} is clearly defined as the new knowledge word, in social debiasing there is generally obscure what is the target to edit out.

\begin{table}[]
\renewcommand{\arraystretch}{1.05} 
  \resizebox{\columnwidth}{!}{%
  \begin{tabular}{ccccccc}
  \toprule
  \multirow{2}{*}{\textbf{Split}} & \multirow{2}{*}{\textbf{Bias type}} & \multirow{2}{*}{\textbf{Number}} & \multicolumn{2}{c}{\textbf{Prompt}} & \multicolumn{2}{c}{\textbf{Target}} \\ \cline{4-7} 
   &  &  & Mean & Std & Mean & Std \\ \midrule
  \multirow{5}{*}{Edit} & All & 929 & 3.27 & 2.46 & 9.70 & 6.24 \\
   & Race & 393 & 3.27 & 2.47 & 8.94 & 5.92 \\
   & Gender & 113 & 3.26 & 2.21 & 9.21 & 6.00 \\
   & Religion & 44 & 3.20 & 2.87 & 11.23 & 5.61 \\
   & Profession & 379 & 3.27 & 2.46 & 10.45 & 6.58 \\ \midrule
  \multirow{5}{*}{Train} & All & 1162 & 3.07 & 2.16 & 9.49 & 6.09 \\
   & Race & 528 & 2.91 & 2.15 & 9.10 & 5.73 \\
   & Gender & 133 & 3.29 & 2.36 & 9.32 & 6.34 \\
   & Religion & 41 & 2.71 & 2.05 & 9.02 & 5.00 \\
   & Profession & 460 & 3.23 & 2.12 & 10.03 & 6.45 \\ \midrule
  \multirow{5}{*}{Val} & All & 232 & 3.09 & 2.06 & 9.69 & 6.17 \\
   & Race & 98 & 2.67 & 1.89 & 9.35 & 6.10 \\
   & Gender & 36 & 3.36 & 1.90 & 10.06 & 6.82 \\
   & Religion & 12 & 3.17 & 2.15 & 8.42 & 4.89 \\
   & Profession & 86 & 3.43 & 2.21 & 10.12 & 6.09 \\ \bottomrule
  \end{tabular}%
  }
  \caption{The statistics of data for LLaMA-2. The data consists of three splits, each split containing four types of bias. Additionally, the average length and standard deviation of prompts and targets are also displayed.}
  \label{table:data_statistics}
\end{table}

% DuNE~\cite{akyurek-etal-2023-dune} propose a benchmark for 
% Internal editing and external editing require different data types.
% 

\begin{table*}[t!]
  \centering
  \small
  \renewcommand{\arraystretch}{1.2}
  \begin{tabular}{cccccc}
  \toprule
  &  &  & \multicolumn{2}{c}{\textbf{Generalization}$\uparrow$} &  \\ \cline{4-5}
    \multirow{-2}{*}{\textbf{Editor}} & \multirow{-2}{*}{\textbf{Success Rate}$\uparrow$} & \multirow{-2}{*}{\textbf{Kownledge Acc}$\uparrow$} & \textbf{$\text{GEN}_{\text{forward}}$} & \textbf{$\text{GEN}_{\text{backward}}$} & \multirow{-2}{*}{\textbf{Average}} \\ \hline
    \multicolumn{1}{c}{Before Edit} & 0.00 & 100.00 & 14.75 & 0.00 & - \\
    \midrule
    \multicolumn{6}{c}{\emph{Internal Editing Algorihtm}} \\
    \midrule
    \multicolumn{1}{c}{FT} & 40.04 & 97.25 & 44.03 & 1.72 & 45.76 \\
    \multicolumn{1}{c}{FT-L} & 3.78 & 98.57 & 20.86 & 0.00 & 30.80 \\
    \multicolumn{1}{c}{MEND} & 91.71 & 96.73 & 81.38 & \textbf{6.24} & 69.01 \\
    \multicolumn{1}{c}{ROME} & 95.80 & 97.38 & 94.62 & 4.95 & 73.19 \\
    \multicolumn{1}{c}{MEMIT} & 94.19 & 98.82 & 88.16 & 3.12 & 71.07 \\
    \midrule
    \multicolumn{6}{c}{\emph{External Editing Algorihtm}} \\
    \midrule
    \multicolumn{1}{c}{SERAC} & 99.25 & \textbf{99.62} & 97.95 & 2.80 & \textbf{74.91} \\
    \multicolumn{1}{c}{IKE} & \textbf{100.0} & 74.32 & \textbf{100.0} & 0.00 & 68.58 \\
  \bottomrule
  \end{tabular}
  \caption{Comparison between editing methods on LLaMA2-7B for single-edit. The best performance is marked in bold. }
  \label{table:single_edit}
  \end{table*}

External editing, which mainly utilizes a retrieval database, generally puts minor requirements over the dataset. 
Taking the whole sentence as the target and not split between target and prompt~\cite{akyurek-etal-2023-dune} suffice the need. 
However, internal editing requires more detailed information such as prompt, target, subject, and paraphrase, as it needs to locate the targeted parameters of the model. 
To facilitate both editing algorithms, we take an approach that requires minimal effort, in order to suit real-world scenarios.  
We construct our dataset based on StereoSet~\cite{stereoset}. 
For each sentence pair, the unbiased sentence is minimally different from the biased one and there is a common subject for both sentences. 
% For the issue of prompt-target split, 
We set the prompt to be the sentence part before the first occurrence of the subject~(including the subject itself). 
The target is the rest of the sentence. 
In this way, less annotation is required and previous datasets with pairing sentences could be utilized. 
For evaluation, we sample sentences from CounterFact~\cite{meng2023locating} to evaluate whether the edit affects LLMs' knowledge. 
We use \texttt{gpt-3.5-turbo-1106}\footnote{\url{https://openai.com/}} to generate three paraphrases for both sentences, to support training methods like MEND and our generalization evaluations. 
An example of our constructed data is shown in Figure \ref{fig:json_data}.
For each model, we only include sentence pairs that exhibit biases (defined in Equation \ref{eq:bias}). 
We split the data into train, valid, and edit sets. 
The statistics of LLaMA-2 can be found in Table \ref{table:data_statistics}.
We can see that our target part is generally longer than the prompt, containing about 10 tokens. Compared to knowledge editing with generally 1-2 tokens as the target, formulation of debias editing poses more challenges to editing algorithms with the need to accurately find the most biased parts in the target.

\subsection{Evaluation Metrics} 
For evaluation, we propose the following four metrics for debias editing. 
\begin{itemize}
  \item \textbf{Edit Success Rate} evaluates whether the likelihood $x_{\text{more}} < x_{\text{less}}$ after edit. This metric indicates the bias level after the edits. 
  \item \textbf{Knowledge Acc} evaluates whether the prediction of LLMs over an unrelated Wikipedia prompt $\hat{p}$ has changed. This metric denotes the locality of editing algorithms. 
  \item \textbf{Generalization} evaluates whether the edited knowledge can generalize to closely related paraphrases. We have two sub-metrics. The first one is $\text{GEN}_{\text{forward}}$, which measures $\inf_j (P(x_{\text{less}}) - P(\hat{x}_{\text{more}}^j)) > 0$. $\hat{x}_{\text{more}}^k$ is the $k$-th paraphrase sentence of $x_{\text{more}}$. It evaluates whether the edited unbiased sentence can beat all paraphrases of the biased sentence. 
  The second metric is called $\text{GEN}_{\text{backward}}$, which evaluates $\inf_{i,j} (P(\hat{x}_{\text{less}}^i) - P(\hat{x}_{\text{more}}^j)) > 0$. It denotes whether the paraphrases of the edited sentence can beat the unedited sentence and its paraphrases. 
  \item \textbf{General Capability} We also evaluate the general capability of LLMs. We choose four benchmarks that are commonly used for evaluating LLMs, including Crows-Pairs~\cite{crows-pairs}, OpenbookQA~\cite{openbookqa}, TruthfulQA~\cite{truthfulqa}, and WinoGrande~\cite{winogrande}. Among these benchmarks, Crows-Pairs evaluates social biases and serves as an out-of-domain test set for our settings. 
\end{itemize}

% - how to define de-biasing with model editing
\section{Experimental Results}
% \subsection{Experimental Setup}

% We refer to our dataset as xxx. 

\begin{figure*}[t]
  \centering
  \includegraphics[width=\linewidth]{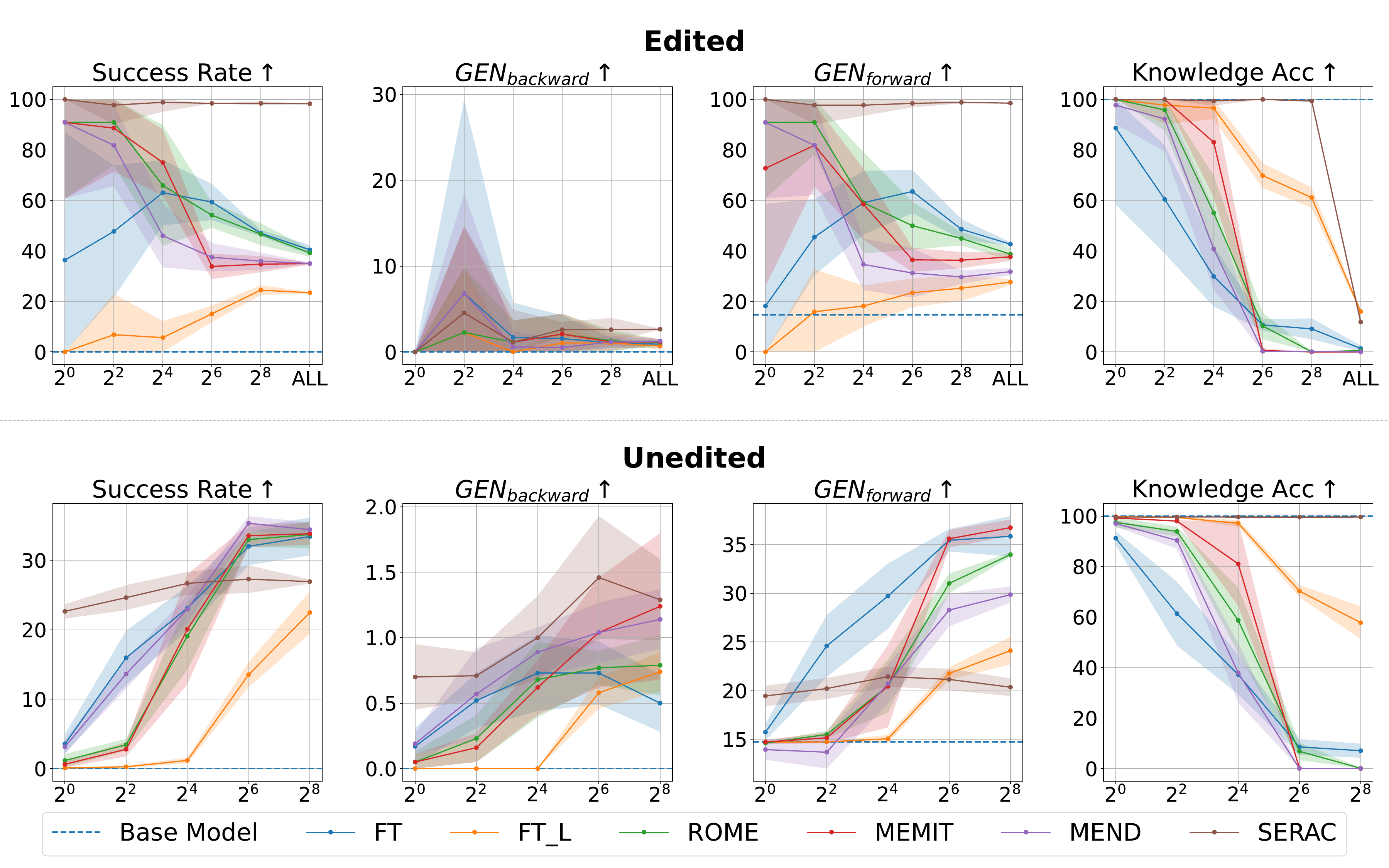}
  \caption{Results of the sequential edit setting. The x-axis is the number of edits performed, and the y-axis is the score. We evaluate both the Edited (Up) and Unedited (Bottom) parts of the data. The gray shadow represents the variance for each method. To reduce variance, we run $[2^0, 2^2, 2^4]$ edits for ten runs and $[2^6, 2^8, \text{ALL}]$ edits for three runs. }
  \label{fig:sequential_edit}
\end{figure*}

% \subsection{Results}
\subsection{Single Edit Results}
Following previous work in model editing~\cite{meng2023locating,meng2023massediting,zhang2024comprehensive}, we first benchmark the editing algorithms under the single edit scenario, in which the LLM is edited with a single bias at a time.
Table \ref{table:single_edit} shows the single edit results with LLaMA2-7B~\cite{touvron2023llama} and results for GPT2-XL can be found in the Appendix. 
In terms of the edit success rate, we can see that \texttt{SERAC} and \texttt{IKE} achieve the strongest performance, with nearly 100\% of the edit rate. \texttt{ROME}, \texttt{MEMIT}, and \texttt{MEND} achieve satisfying scores of $>90\%$ edit rate. \texttt{FT} and \texttt{FT-L} are the worst, which cannot reliably perform a single edit. 
These results demonstrate the great potential of editing methods to override stereotypical biases. 
% \texttt{FT} and \texttt{FT-L} achieve the worst edit rate, about 40\%. 

For the Knowledge Acc, which denotes whether the editing algorithms can accurately edit the bias without affecting LLMs' knowledge, we find that most methods have high Knowledge Acc~($>97\%$), showing that they can effectively edit out bias without hurting LLM's internal knowledge. 
One exception is IKE, which only obtains a score of 74.32\%. 

Furthermore, we evaluate the generalization of these editing methods over debiasing scenarios. We find that the editing methods can effectively improve the target sentence, denoting by the high $\text{GEN}_{\text{forward}}$, but they struggle with the $\text{GEN}_{\text{backward}}$. Recall that $\text{GEN}_{\text{backward}}$ measures whether the paraphrases of the unbiased sentence can beat biased ones. 
These results suggest that one of the challenges of model editing methods is to generalize over semantically equivalent sentences. 

\begin{table*}[t]
\centering
\small
\renewcommand{\arraystretch}{1.3}
\begin{tabular}{ccccc}
\toprule
\multicolumn{1}{l}{} & \textbf{Crows-Pairs}$\downarrow$ & \textbf{OpenbookQA}$\uparrow$ & \textbf{WinoGrande}$\uparrow$ & \textbf{TruthfulQA}$\uparrow$ \\ \midrule
Before Edit & 66.91        & 31.4         & 69.14        & 38.96        \\ \midrule
% \textbf{LoRA} & 45.22 ± 0.40 & 15.93 ± 1.80	& 49.37 ± 0.27 & 50.91 ± 1.04 \\ \hline
FT          & 47.92 ± 0.09 & 13.13 ± 1.62 & 49.51 ± 0.73 & 48.98 ± 1.09 \\
FT-L       & 64.78 ± 0.81 & \textbf{31.33} ± 0.31 & \textbf{68.56} ± 0.41 & 39.14 ± 0.22 \\
MEND        & 52.69 ± 1.38 & 15.20 ± 1.40 & 50.72 ± 2.40 & 49.14 ± 1.62 \\
SERAC      & 54.46 ± 0.03 &	\underline{17.60} ± 0.00 &38.74 ± 0.00	& \textbf{51.93} ± 0.00 \\
MEMIT       & \textbf{42.12} ± 3.41 & 14.87 ± 1.50 & \underline{50.75} ± 1.71 & 48.55 ± 1.31 \\
ROME        & \underline{47.47} ± 1.89 & 14.53 ± 1.30 & 50.07 ± 0.37 & \underline{50.04} ± 0.74 \\ \bottomrule
\end{tabular}%
\caption{Comparison of the performance for four benchmarks between before and after edits. The best performance is marked in bold and the second performance is marked with underline. All mean and variance are reported across three runs.}
\label{table:harness_results}
\end{table*}

\subsection{Sequential Edit}
\label{sec:seq_edit}
Another important scenario is the sequential edit, where the LMs are edited with a stream of requests. 
The scenario is closely aligned with debiasing in real-world applications where biased sentences generated by LLMs (bad cases) are found one by one. 
% This setting is more challenging as the edited requests for each LLMs
% In this setting, we expect the model to achieve a good edit success rate as well as 

The results are shown in Figure \ref{fig:sequential_edit}. 
We do not include \texttt{IKE} in this setting as \texttt{IKE} only supports single-edit. 
During the sequential process, we evaluate both the already edited and non-edited data for this setting. We have the following observations: 
\begin{itemize}
    \item The edit success rate decreases with the increase of edits. After hundreds of edits, the success rate of already-edited requests gradually converges to about 40\%, except for SERAC, which achieves nearly 100\% success rate. In addition, we find that the unedited part of the data also improves with edits, indicating that editing algorithms achieve a certain extent of generalization in solving biases.
    \item For all methods, the Knowledge ACC retains high with a small number of edits but quickly decreases when the number of edits gets larger. FT affects Knowledge ACC the most, as the score decreases from 100 even with only 1 edit. Other methods like \texttt{ROME}, \texttt{MEMIT}, and \texttt{MEND} suffer less from this problem but still demonstrate a significant performance drop with more than 32 edits. \texttt{FT-L} and \texttt{SERAC} are the most robust against the growth of edits. These results suggest a key challenge of editing in debiasing is to keep the edits specific and accurate with a large number of edits. 
    \item For generalization metrics, we find that editing methods generally perform well on $\text{GEN}_{\text{forward}}$, in both edited or unedited evaluations. However, $\text{GEN}_{\text{backward}}$ is more challenging, as discussed in the previous section. Hence, another key challenge we identified for debias editing is to generalize over targets.
    \item For all edited evaluations, we observe a strong performance of \texttt{SERAC}, achieving a high success rate and having the smallest influence on the model's internal knowledge, which suggests utilizing memories and retrieval is a promising direction for debias editing. Nevertheless, we also observe drawbacks of \texttt{SERAC} when generalizing to unedited data. 
\end{itemize}
% It is a promising result compared with the baseline model, but there is still a lot of room to improve. 
% \subparagraph*{Edit Results}
% Figure \ref{} presents the curve of metrics along with the edit numbers.

\subparagraph*{General Capability}
In this section, we compare the general capability of LLMs before and after edits. Results are shown in Table \ref{table:harness_results}, and we use \emph{lm-harness}~\cite{eval-harness} for evaluation. First of all, we find that after edits, the Crows-Pairs, which are commonly used to evaluate the bias of LLMs, demonstrates a lower level of biases for all editing algorithms. \texttt{MEMIT} performs the best across all editing methods, reducing the bias score from 66.91 to 42.12. 

Then, for OpenbookQA and WinoGrande~(benchmarks for world knowledge and reasoning ability), editing methods lead to clear drops in performance, showing a tradeoff between helpfulness and bias mitigation. \texttt{FT-L} achieves the minimum decline in OpenbookQA and WinoGrande, but also the smallest improvement on Crows-Pairs. 

Interestingly, we observe the model's performance on TruthfulQA increases after edits, ranging from 1\% to 11\%, suggesting that solving stereotypical biases makes the model more truthful. We conjecture that the representation of truthfulness and bias~\cite{zou2023representation} inside LLMs might be connected.

\begin{table}[h]
\centering
\renewcommand{\arraystretch}{1.2} 
\resizebox{\columnwidth}{!}{%
\begin{tabular}{cccccc}
\toprule
\multirow{2}{*}{\textbf{Edit}} & \multirow{2}{*}{\textbf{Algorithm}} & \multicolumn{4}{c}{\textbf{Eval}} \\ \cline{3-6} 
 &  & \textbf{Race} & \textbf{Gender} & \textbf{Religion} & \textbf{Profession} \\ \hline
\multirow{6}{*}{Race} & FT & 43.26 & \underline{32.74} & 27.27 & 35.36 \\
 & FT-L & 24.68 & 15.93 & 11.36 & 21.90 \\
 & MEND & 32.82 & \textbf{33.63} & \underline{27.27} & \textbf{36.41} \\
 & SERAC & \textbf{99.24} & 23.01 & \textbf{34.09} & 28.23 \\
 & ROME & \underline{45.80} & \textbf{33.63} & 25.00 & 34.04 \\
 & MEMIT & 35.11 & 30.09 & \underline{27.27} & \underline{35.62} \\ \hline
\multirow{6}{*}{Gender} & FT & 33.08 & \underline{48.67} & 22.73 & \underline{34.56} \\
 & FT-L & 11.70 & 23.01 & 0.00 & 17.41 \\
 & MEND & 33.33 & 37.17 & 22.73 & \textbf{35.62} \\
 & SERAC & 23.16 & \textbf{98.23} & 20.45 & 28.23 \\
 & ROME & \textbf{36.90} & 42.48 & \textbf{27.27} & 30.87 \\
 & MEMIT & \underline{35.88} & 33.63 & \underline{25.00} & 32.98 \\ \hline
\multirow{6}{*}{Religion} & FT & 33.08 & 25.66 & \underline{38.64} & 28.23 \\
 & FT-L & 2.29 & 7.08 & 11.36 & 1.85 \\
 & MEND & 33.59 & \underline{33.63} & 18.18 & 36.41 \\
 & SERAC & 28.50 & 24.78 & \textbf{100.00} & 27.97 \\
 & ROME & \textbf{35.37} & \underline{33.63} & 36.36 & \textbf{38.52} \\
 & MEMIT & \underline{34.10} & \textbf{35.40} & 31.82 & \underline{36.15} \\ \hline
\multirow{6}{*}{Profession} & FT & 32.82 & \underline{31.86} & 22.73 & 42.48 \\
 & FT-L & 13.99 & 15.93 & 6.82 & 20.05 \\
 & MEND & 33.59 & 29.20 & \underline{25.00} & 34.30 \\
 & SERAC & 27.48 & \underline{31.86} & \textbf{27.27} & \textbf{97.89} \\
 & ROME & \textbf{37.91} & \textbf{32.74} & 20.45 & \underline{44.06} \\
 & MEMIT & \underline{35.88} & \textbf{32.74} & 25.00 & 33.25 \\ \bottomrule
\end{tabular}%
}
\caption{The generalization among different bias types. The best performance is marked in bold, and the second to best is marked in underline. }
\label{table:bias_type_results}
\end{table}

\begin{table*}[t!]
\centering
\small
\renewcommand{\arraystretch}{1.3}
\begin{tabular}{cccccccc}
\hline
\textbf{Metric} & \textbf{Method} & \textbf{1} & \textbf{4} & \textbf{16} & \textbf{64} & \textbf{256} & \textbf{ALL} \\ \hline
\multirow{3}{*}{Success Rate} & Baseline & 36.36 ± 50.45 & 47.73 ± 26.11 & 63.07 ± 12.95 & 59.37 ± 7.16 & 47.00 ± 1.26 & 40.44 ± 1.94 \\
 & Causal & 45.45 ± 52.22 & 43.18 ± 19.66 & 52.84 ± 10.59 & 54.17 ± 7.86 & 53.52 ± 1.41 & 53.18 ± 0.67 \\
 & Rule & \textbf{63.64} ± 36.36 & \textbf{79.55} ± 20.45 & \textbf{71.59} ± 10.96 & \textbf{78.65} ± 5.49 & \textbf{78.65} ± 0.23 & \textbf{71.13} ± 1.53 \\ \hline
\multirow{3}{*}{$\text{GEN}_{\text{backward}}$} & Baseline & 0.00 ± 0.00 & \textbf{6.82} ± 22.61 & 1.70 ± 4.04 & 1.56 ± 2.71 & 1.04 ± 0.60 & 0.90 ± 0.38 \\
 & Causal & 0.00 ± 0.00 & 0.00 ± 0.00 & \textbf{2.84} ± 5.84 & 3.12 ± 1.57 & 1.56 ± 0.39 & 1.58 ± 0.12 \\
 & Rule & 0.00 ± 0.00 & 2.27 ± 7.54 & 1.70 ± 4.04 & \textbf{4.17} ± 4.78 & \textbf{3.65} ± 0.82 & \textbf{2.91} ± 0.47 \\ \hline
\multirow{3}{*}{$\text{GEN}_{\text{forward}}$} & Baseline & 18.18 ± 40.45 & 45.45 ± 15.08 & \textbf{59.09} ± 12.61 & \textbf{63.54} ± 8.61 & 48.57 ± 4.01 & 42.67 ± 0.81 \\
 & Causal & \textbf{27.27} ± 46.71 & 31.82 ± 22.61 & 43.75 ± ~7.91 & 43.75 ± 11.27 & 50.39 ± 0.39 & 50.34 ± 1.15 \\
 & Rule & \textbf{27.27} ± 46.71 & \textbf{54.55} ± 36.77 & 55.11 ± 14.47 & 56.77 ± 8.02 & \textbf{59.77} ± 4.11 & \textbf{56.99} ± 1.62 \\ \hline
\multirow{3}{*}{Knowledge} & Baseline & \textbf{88.64} ± 11.36 & \textbf{60.42} ± 21.47 & 29.88 ± 11.89 & 10.68 ± 2.27 & 9.14 ± 4.15 & 1.36 ± 1.18 \\
 & Causal & 88.64 ± 11.36 & \textbf{64.02} ± 29.17 & 28.46 ± 14.78 & 13.54 ± 3.76 & 8.29 ± 4.00 & 5.67 ± 2.13 \\
 & Rule & 81.82 ± 18.18 & 51.70 ± 29.83 & \textbf{33.90} ± 17.32 & \textbf{24.18} ± 8.55 & \textbf{12.07} ± 1.88 & \textbf{8.86} ± 4.12 \\ \hline
\end{tabular}%
\caption{Experimental results for Causal tracing and Rule-based methods with FT as our baseline. The x-axis is the number of samples used in Sequential-Edit. }
\label{tab:method}
\end{table*}

\subsection{Generalization over Bias Types}
Furthermore, we wonder if edits of a certain bias, e.g., Race, can be generalized to another unseen type of bias, e.g., Profession. 
We consider four bias types, Race, Gender, Religion, and Profession. 

As shown in Table \ref{table:bias_type_results}, we perform edits on one bias type and evaluate on all four bias types. 
In general, editing methods demonstrate generalization across bias types. \texttt{SERAC} performs the best among in-domain edits (e.g., edit on Race and evaluate on Race), and \texttt{ROME} and \texttt{MEMIT} perform the best for generalization. Most of the methods achieve about ~30\% of success rate when edits transfer, further demonstrating the potential of using editing methods to solve bias in LLMs. 

\section{Methods}
As mentioned before, one key challenge we identified for debias editing is to accurately locate the biased part in the target and make minimal modifications. 
Thus, we propose two simple methods to address this issue.

\paragraph{Rule-based}
We first propose a heuristic rule that selects the most informative parts of the sentence.  
% filters the target phrase with . 
We hypothesize that tokens with more informative part-of-speech~(POS-Tagging, \citealt{stanza}) tags are more valuable targets for reversing the biases in LLMs. We use spacy~\footnote{\url{https://spacy.io/}} to parse each sentence and filter out ["DET', "AUX", "PUNCT", "PRON", "ADP"]. Furthermore, we compare the biased and unbiased pairs in our dataset that have minimal lexical differences and extract the longest common prefix of two sentences. 

\paragraph{Causal Tracing}
Our second method utilizes causal tracing to find the most informative parts that correlate to the subject. 
In particular, we first measure baseline
predicted probabilities of each target token when noise is introduced into the encoding of the subject tokens
to degrade the effect of the subject for the model. As in \cite{meng2022locating}, we use Gaussian noise with standard deviation $3\sigma$ ($\sigma ^2$ is the empirically observed variance of embedding activations). Then we take the top 5 tokens with the highest probability reduction as the new targets.

\paragraph{Experimental Results}
The experimental results for both methods can be found in Table \ref{tab:method}. We use FT as our baseline. 
In terms of the Success Rate, we can see that both methods improve the scores with the increase of samples edited. Our rule-based method consistently surpasses causal and the FT baseline. 
As for Generalization scores, we find that even with our methods, the models still suffer from generalization backward. On the plus side, our methods improve $\text{GEN}_{\text{forward}}$ with `256' and `ALL' samples.
For Knowledge ACC, we observe that the rule-based method can consistently retain more knowledge even after 929 edits. 
These results demonstrate the effectiveness of our two methods and verify our intuition that the selection of parts to edit out is a key challenge for debias editing. 

% Please add the following required packages to your document preamble:
% \usepackage{multirow}
% \usepackage{graphicx}

\section{Conclusion}
% \elliott{rewrite}
In this paper, we comprehensively study the problem of model editing for stereotypical debiasing. After designing viable formulation that supports both internal and external editing algorithms, we evaluate seven model editing methods and observe the potential and challenges for debias editing in the single edit, sequential edit, and bias generalization. 
Furthermore, based on our observations, we propose two simple methods to address the challenge we find, demonstrating promising results. 

\section{Limitations}
% We have identified the challenge of generalizing debias edit effect to paraphrases. However, in this paper, we do not propose a method that can address this challenge.  

% dataset

% implicit bias
Currently, we only focus on the explicit biases present in sentences, while the definition, evaluation, and mitigation of implicit biases have not been considered.
% more scenarios
We have not considered additional scenarios where biases may exist and manifest, such as in chat or question answering.
% language
We are currently working exclusively with English-language datasets and have not considered other languages.
We mainly focus on foundation models, where chat models are not considered in our settings. 

\section{Ethical Considerations}
This paper focuses on debiasing editing. However, the same technique can be applied to enhance the bias of LLMs. The fair use of editing techniques needs more attention. 
Our scope was limited by foundational datasets such as StereoSet, which means that not all ethnicities, politics, or religious perspectives were included.
We honor the ACL Code of Ethics. No private data or non-public information was used in this work. 

\bibliography{anthology,custom}
\bibliographystyle{acl_natbib}
% Custom bibliography entries only
% \bibliography{custom}

\clearpage
\appendix

% \section{Example Appendix}
\section{Experimental Details}
\label{sec:exp_detail}

In this section, we discuss more about the experimental details. 
% \elliott{discuss layer selection for FT, FT-L, ROME, and MEMIT}
The layer used for FT, FT-L, ROME, and MEMIT follows previous work~\cite{meng2023locating,meng2023massediting,wang2023easyedit}.
Specifically, for LLaMA-2, FT and ROME use the 5-th layer, and for GPT2-XL, FT and ROME use the 17-th layer.

All experiments are conducted with NVIDIA A100 GPUs. A single edit experiment takes about 8 GPU hours for LLaMA-2 and about 2 GPU hours for GPT2-XL. 
% For FT and FT-L, we use the 5-th layer for LLaMA-2 and the 0-th layer for GPT2-XL, based on the results of causal tracing~\cite{meng2023locating}. 

\section{GPT2-XL Results}
\subsection{Single Edit}
In Table \ref{table:single_edit_gpt2} we present the single edit results for GPT2-XL. We can see that, methods like SERAC and IKE achieve high success rates, and ROME balances well between success rate and Knowledge ACC. 
Methods like MEND and FT-L fail to edit out biases for GPT2-XL, which is not observed with LLaMA-2. 
We conjecture the reason is the smaller size of GPT2-XL brings more challenges to balancing the effect of performing success edits and less harm to the models' knowledge. 

\section{Details for Data Preparations}
\subsection{Data examples}
We present a JSON-format example of our data, as shown in Figure \ref{fig:json_data}. 

\subsection{Prompt for paraphrases}
\begin{figure}[t]
  \centering
  \includegraphics[width=1.0\columnwidth]{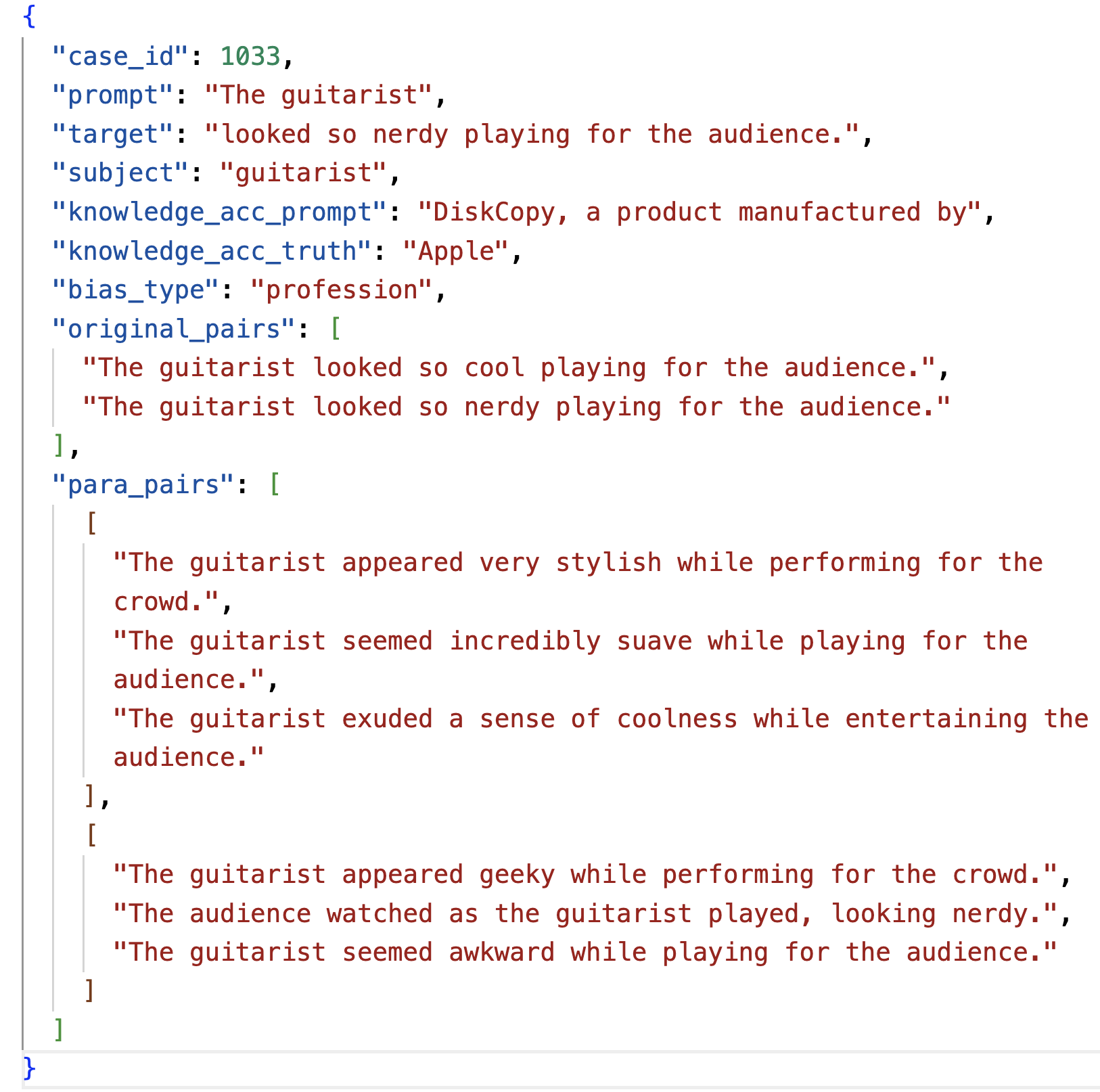}
  \caption{An example of the data structure in JSON format.}
  \label{fig:json_data}
\end{figure}

We use the following prompt to generate paraphrases for both biased and unbiased sentences. 
\begin{quote}
    Can you help me paraphrase the following sentence. Please give me three candidate paraphrases, and put each paraphrase in one line. Make sure to keep the word \{SUBJECT\}.
\end{quote}
\{SUBJECT\} is the subject of that biased sentence. For example, \{SUBJECT\} is `guitarist' in the example of Figure \ref{fig:json_data}

\subsection{Sequential Edit}
\begin{figure*}[t]
  \centering
  \includegraphics[width=\linewidth]{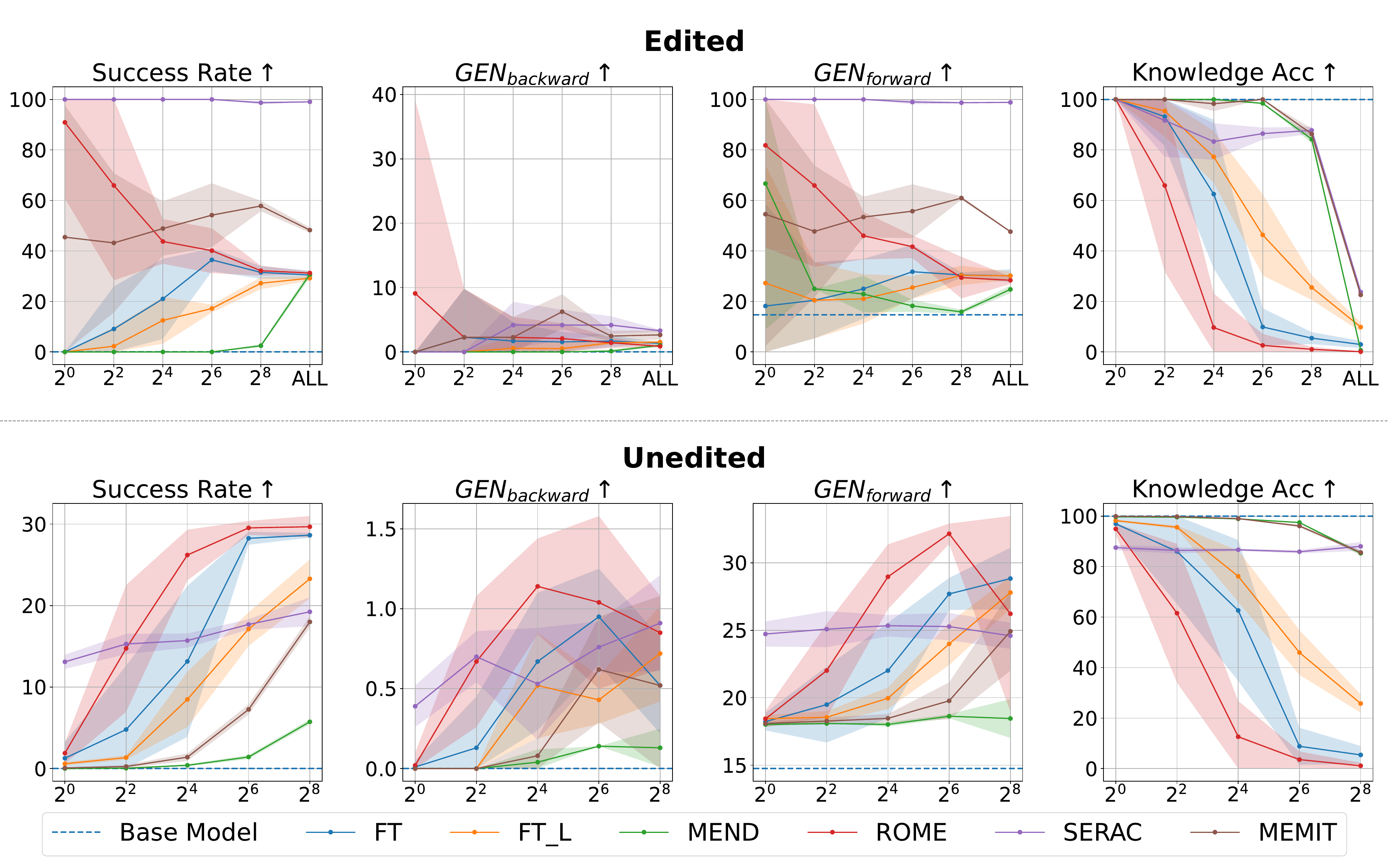}
  \caption{Results of the sequential edit setting on GPT2-XL. The x-axis is the number of edits performed, and the y-axis is the score. We evaluate both the Edited (Up) and Unedited (Bottom) parts of the data. The gray shadow represents the variance for each method. To reduce variance, we run $[2^0, 2^2, 2^4]$ edits for ten runs and $[2^6, 2^8, \text{ALL}]$ edits for three runs. }
  \label{fig:sequential_edit_gpt2}
\end{figure*}

We also conduct experiments with GPT2-XL on Sequential Edit. As shown in Figure \ref{fig:sequential_edit_gpt2}, we observe a similar trend as in the main content and the conclusions are the same as in Section \ref{sec:seq_edit}.

\begin{table*}[t!]
\renewcommand{\arraystretch}{1.15} 
  \centering
  \small
  \renewcommand{\arraystretch}{1.3}
  \begin{tabular}{cccccc}
  \toprule
  &  &  & \multicolumn{2}{c}{\textbf{Generalization}$\uparrow$} &  \\ \cline{4-5}
    \multirow{-2}{*}{\textbf{Editor}} & \multirow{-2}{*}{\textbf{Success Rate}$\uparrow$} & \multirow{-2}{*}{\textbf{Kownledge Acc}$\uparrow$} & \textbf{$\text{GEN}_{\text{forward}}$} & \textbf{$\text{GEN}_{\text{backward}}$} & \multirow{-2}{*}{\textbf{Average}} \\ \hline
    \multicolumn{1}{c}{Before Edit} & 0.00 & 100.00 & - & - & - \\
    \midrule
    \multicolumn{6}{c}{\emph{Internal Editing Algorihtm}} \\
    \midrule
    FT          & 15.91  & 93.09  & 24.90 & 0.39 & 33.57                \\
    FT-L        & 3.78   & 98.57  & 20.86 & 0.00 & 30.80                \\
    MEND        & 0.00   & 99.87  & 16.04 & 0.00 & 28.98                \\
    ROME        & 82.27  & 90.48  & 80.31 & 3.91 & 64.24                \\
    MEMIT       & 39.77  & 99.74  & 48.37 & 2.74 & 47.65                \\
    \midrule
    \multicolumn{6}{c}{\emph{External Editing Algorihtm}} \\
    \midrule
    SERAC       & 100.00 & 87.35  & 99.48 & 3.26 & 72.52                \\
    IKE         & 100.00 & 68.97  & 98.31 & 2.61 & 67.47                \\
  \bottomrule
  \end{tabular}
  \caption{Comparison between editing methods on GPT2-XL for single-edit. The best performance is marked in bold. }
  \label{table:single_edit_gpt2}
  \end{table*}

\end{document}